\documentclass[12pt]{article}
\usepackage{amsmath,amsthm}
\usepackage{graphicx}
\usepackage{hyperref}
\usepackage{epsfig}
\usepackage{psfrag}
\usepackage{graphics}
\usepackage{setspace}
\usepackage{anysize}

\marginsize{1.25in}{1.25in}{1.0in}{1.375in}

\begin{document}
\title{\textbf{Concept of E-machine: \\How does a "dynamical" brain learn\\ to process "symbolic"
information? } \\\textbf{Part I} \footnote{Accepted for
publication in "Progress in Computer Science Research", \emph{Nova
Science Publishers, Inc}. }}
\author{Victor Eliashberg \\
\\
\emph{Stanford University, Department of Electrical
Engineering}\\
web: www.brain0.com, email: victor@brain0.com}
\date{}
\maketitle
\begin{abstract}
The human brain has many remarkable information processing
characteristics that deeply puzzle scientists and engineers. Among
the most important and the most intriguing of these
characteristics are the brain's broad universality as a learning
system and its mysterious ability to dynamically change
(reconfigure) its behavior depending on a combinatorial number of
different contexts.

This paper discusses a class of hypothetically brain-like
dynamically reconfigurable associative learning systems that shed
light on the possible nature of these brain's properties. The
systems are arranged on the general principle  referred to as the
\emph{concept of E-machine}.

The paper addresses the following questions:
\begin{enumerate}
\item \emph{How can "dynamical" neural networks function as
universal programmable "symbolic" machines?} \item \emph{What kind
of a universal programmable symbolic machine can form arbitrarily
complex software in the process of programming similar to the
process of biological associative learning?} \item \emph{How can a
universal learning machine dynamically reconfigure its software
depending on a combinatorial number of possible contexts?}
\end{enumerate}

The paper explains the concept of E-machine and outlines a broad
range of its potential applications. These applications include:
context-sensitive associative memory, context-dependent pattern
classification, context-dependent motor control, imitation,
simulation of complex "informal" environments and natural
language.
\end{abstract}
 \setcounter{section}{-1}
\section{Introduction} \label{sec0}

When observed "from the outside" the human brain seems to behave
as a sequential symbolic machine. How else can one explain such
"clearly symbolic" phenomena as mental computations and natural
language? When observed "from the inside," however, the neural
networks of the brain evoke an idea of a noisy dynamical system
with distributed parameters rather than the image of a logic
circuitry of a digital computer -- gradually changing potentials,
decaying residual excitation, high level of fluctuations. Neurons
do produce spikes reminiscent of the pulses in a digital computer.
It is widely believed, however, that it is the frequency of these
pulses rather than their presence and absence that carry the
important information.

\begin{enumerate}
\item \emph{How can "dynamical" neural networks function as
universal programmable "symbolic" machines?} \item \emph{What kind
of a universal programmable symbolic machine can form arbitrarily
complex software in the process of programming similar to the
process of biological associative learning?} \item \emph{How can a
universal learning machine dynamically reconfigure its software
depending on a combinatorial number of possible contexts?}
\end{enumerate}

The metaphor "the brain as an E-machine" (Eliashberg, 1967, 1979,
1981, 1989, 1990b) sheds light on these questions. The metaphor
suggests that the brain is neither a traditional symbolic system,
nor is it a traditional dynamical system. It is a "non-classical
symbolic system" in which the probabilities of sequential discrete
("symbolic") processes are controlled by the massively parallel
continuous ("dynamical") processes.\\\\
\textbf{Note}. The general idea that the brain employs a
combination of symbolic and dynamical computational mechanisms was
entertained in different forms by different researchers (Collins
and Quillian, 1972; Anderson, 1976; and many others.) The concept
of E-machine is an attempt to provide a neurobiologically
consistent formalization of this general idea. The requirement of
neurobiological consistency makes a big difference!\\\\
 The paper is divided into two parts. Part I consists of the
three main sections:
\begin{enumerate}
\item \textbf{The Whole Human Brain as a Universal Learning
Computer}. This section takes a broader look at the problem of
information processing in the whole human brain. It argues that
there exists a relatively short formal representation of a
universal learning computer similar to an untrained (unprogrammed)
human brain.

\item \textbf{From Associative Neural Networks to E-machines}.
This section establishes a  link between associative neural
networks and E-machines. It connects the effects of dynamic
reconfiguration (neuromodulation) in neural networks with the
hypothetical states of dynamical memory available in individual
neurons. These states of "residual-excitation-like" memory are
referred to as the E-states.

\item \textbf{Molecular Interpretation of E-states: Ensembles of
Protein Nanomachines as Statistical Mixed-signal Computers}. This
section addresses the problem of a neurobiological implementation
of the E-sates and the next E-state procedures. It describes a
formalism that connects the dynamics of macroscopic E-states with
the statistical conformational dynamics of ensembles of protein
molecules (such as ion channels) embedded in neural membranes. A
single protein molecule is treated as a probabilistic nanomachine,
and the E-states are interpreted as the average numbers of such
nanomachines in different states -- the average occupation
numbers. The formalism suggests that it is the statistical
conformational dynamics of protein molecules in \emph{individual}
neurons rather than the \emph{collective} statistical dynamics of
neural networks that performs the main volume of the brain
hardware computations. There is  not enough neurons in the whole
human brain to implement the required amount of computations in
the networks built from "simple neurons."
\end{enumerate}

Part II includes the following main sections:

\begin{enumerate}
\setcounter{enumi}{3} \item \textbf{Computing with E-states}. This
section tackles the question as to how the massively parallel
transformations of E-states allow a slow brain to efficiently
process  large arrays of symbolic data stored in its long-term
memory (LTM) without moving this data into a read/write memory
buffer.

\item \textbf{Hierarchical structure: sparse-recoding, data
compression and statistical filtering}. This section explains how
E-machines with hierarchical structure of associative memory can
perform efficient data compression, context-dependent statistical
filtering, and context-dependent generalization. \item
\textbf{Discussion}

\end{enumerate}

\section{The Whole Human Brain as a Universal Learning
Computer} \label{sec1}  This section takes a broader look at the
problem of information processing in the whole human brain. It
argues that there exists a relatively short formal representation
of a universal learning computer similar to an untrained
(unprogrammed) human brain.

\subsection{System (Man,World) as a composition of two "machines"}
\label{sec1.1}

\begin{figure}[b!]
\begin{center}
 \includegraphics[width=4.0in]{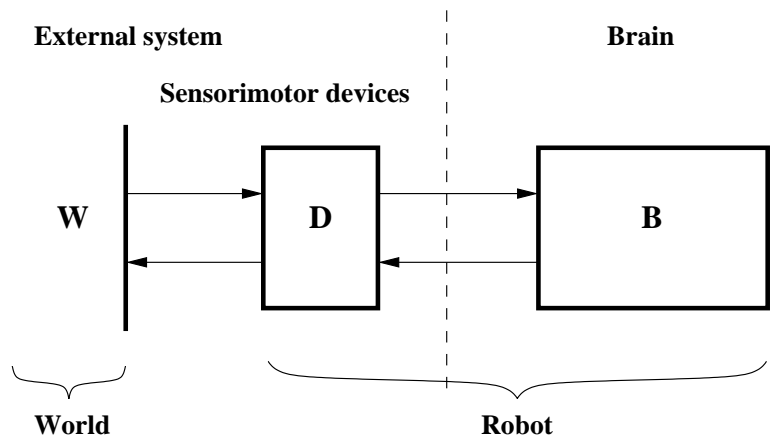}
 \end{center}
 \caption{System (Robot,World) as a composition of two machines} \label{fi1}
\end{figure}

Consider a cognitive system (W,D,B) schematically shown in
Figure~\ref{fi1}, where W is an external world, D is a set of
human-like sensory and motor devices, and B is a hypothetical
computing system simulating the work of the human nervous system.
One can think of system (D,B) as a human-like robot.
From the system-theoretical viewpoint, it is useful to divide
system (W,D,B) into two subsystems: (W,D) and B, where (W,D) is
the external world as it appears to the brain B via devices D. In
this representation, both subsystems can be treated as abstract
"machines", the inputs of B being the outputs of (W,D) and vice
versa.

For the sake of simplicity, I refer to B as the brain. At this
general level, the rest of the nervous system can be treated as a
part of block D. Let B(t) denote the state of B at time t, where
t=0 corresponds to the beginning of learning. I argue that the
following general propositions are true:

\begin{enumerate}
\item There exists a relatively short  formal representation of
B(0). This representation is encoded in the human genome and can
be small enough to fit into a single floppy disk.
 \item No special mathematical formalism is needed to
 describe the work of B(0). Given a powerful enough hardware, a
 relatively small \emph{C} or \emph{C++} program would be able to
 simulate the work
of B(0) with a time step of, say, $1msec$. This program would be
sufficient to adequately represent all important psychological
characteristics of B(0). A more complex, but still rather small
\emph{C} or \emph{C++} program would be able to simulate the work
of B(0) with a time step of, say, $1\mu sec$. This program would
be sufficient to adequately represent all important psychological
characteristics of B(0) and many of its neurobiological
characteristics. \item There exists a relatively short formal
representation of the sensorimotor devices, D, since this
representation is encoded in the human genome. The
\emph{metaphorical floppy disk} mentioned in item 1 has enough
room for both B(0) and D. We know that B(0) can do well with
different kinds of artificial devices, so the main secret is in
B(0) rather than in D. \item In the general case, there exists no
finite formal representation of system (W,D) -- this system can be
infinitely complex. This doesn't prevent one from simulating the
behavior of system (W,D,B), because the "robot" (D,B) has a finite
formal representation, and the external world, W, is "always
there" to experiment with. \item Any formal representation of B(t)
for a big t (say, t$>$10 years) must be very long (terabytes?) --
this representation must include in some form a representation of
the brain's individual experience which resulted from interaction
with (W,D). Whatever language is used for the representation of
B(t), the main part of this representation is the representation
of the knowledge accumulated in the course of learning.
Figuratively speaking, the human brain works as a "complexity
sucker" that gets most if its complexity from system (W,D). \item
The knowledge is represented in B(t) in a rather "raw" form -- the
brain's learning algorithm is close to "memorizing raw
sensory-motor-emotional experience."  No special data structures
are needed. Instead of pre-processing data before putting it in
memory, the brain uses a powerful massively parallel
decision-making procedure capable of processing the "raw"
experience on the fly depending on context. \item It is
practically impossible to understand B(t) without understanding
B(0) and studying the process of learning that changes B(0) into
B(t).
 \item It is practically impossible to formally represent and simulate
 nontrivial parts of the behavior
of system (W,D,B(t)) without having an adequate formal
representation of B(t). That is, an adequate cognitive theory
cannot be separated from the theory of the brain. \item The main
goal of brain modelling must be reverse engineering B(0). This is
a clearly defined and practically achievable goal. (I refer to
this reverse engineering project as the Brain Zero or the Brain 0
project. Visit www.brain0.com.) To advance toward this goal one
should concentrate on the analysis of basic psychological and
neurobiological observations rather than on the mimicking of the
parts of the brain's behavior. The latter strategy leads one into
the "new-effect-new-model" pitfall and is cursed by the
\emph{combinatorial explosion} of the number of partial models
needed to represent the whole behavior. \item The role of B(0) in
cognitive science can be meaningfully compared with the role of
the Maxwell equations in the classical electrodynamics. The same
Maxwell equations (a metaphorical counterpart of B(0)) coupled
with an infinite variety of specific external constraints (a
metaphorical counterpart of (W,D)) allow one to simulate infinite
variety of specific classical electromagnetic phenomena.
Similarly, the same B(0) interacting with different external
systems (W,D) would allow one to simulate, in principle, infinite
variety of arbitrarily complex cognitive phenomena.

\end{enumerate}
\subsection{The Maxwell equations metaphor: the pitfall of a "pure phenomenology"} \label{sect1.2}

The example of physics warns us that one should not underestimate
the power of simple basic mechanisms of Mother Nature. I argue
that this warning is relevant to the problem of reverse
engineering the "physical" system B(0). The brain is designed by
Mother Nature -- not by the human system engineers. This makes all
the difference in the world.

We (humans) design artificial information processing systems to
make them easier to understand, test and debug. This costs us
extra resources. In contrast, Mother Nature tends to solve natural
design problems with minimum resources. It makes Her designs look
clever. It also makes them difficult to understand. In such
minimum-resource designs different functions are necessarily
strongly integrated and cannot be easily structured as independent
blocks.

An integration of a set of simple physical principles can produce
a "critical mass" effect. The introduction of the so-called
"displacement current" in the Maxwell equations gives a classical
example of this interesting phenomenon. All of a sudden, this
simple addition to the set of known basic laws of electricity and
magnetism, allowed J.C. Maxwell to create his famous equations
that cover the whole range of arbitrarily complex classical
electromagnetic phenomena.

I argue that something similar had happened in the case of the
human brain. Not too much was needed to transform the brains of
simple animals into the human brain. A clever integration of a
relatively small set of powerful "basic mechanisms" produced a
"critical mass" effect.

To understand the pitfall of a "pure phenomenology" consider the
following metaphor. Imagine a physicist who wants to simulate the
behavior of electromagnetic field in a complex microwave device,
e.g., the Stanford Linear Accelerator (SLAC). Assume that this
physicist doesn't know about the existence of the Maxwell
equations and, even more importantly, doesn't believe that the
complex behavior he observes may have something to do with such
simple equations. (In the AI jargon this physicist would be called
"scruffy." If he believed in the existence of the basic equations
he would be called "neat.")

So this "scruffy physicist" sets out to do a purely
phenomenological computer simulation of the observed complex
behavior per se. Anyone who was involved in the computer
simulation of the behavior of electromagnetic field in a linear
accelerator can easily predict the results of this gedanken
experiment.

In the best case scenario, the above mentioned scruffy physicist
comes up with a computer program (with a large number of empirical
parameters) capable of simulating the behavior of electromagnetic
field in a very narrow range. This computer program has no
extrapolating power and is not accepted by the SLAC community as a
theory of a linear accelerator.

Note that it would be impossible to reverse engineer the Maxwell
equations (a metaphorical counterpart of B(0)) from the analysis
of the behavior of electromagnetic field in such a complex
"external world" as SLAC. I argue that, similarly, it is
impossible to reverse engineer B(0) from the analysis of such
complex cognitive phenomena in system (W,D,B(t)) as playing chess,
solving complex mathematical problems, story telling, etc.

\subsection{Basic observations }
\label{sec1.3}

To formulate some "technical requirements" to an adequate model of
B(0)  consider the following basic observations:\\\\
\textbf{OBSERVATION 1.} A person with a sufficiently large
external memory aid (for example, a sheet of paper divided into
squares) can perform, in principle, any \emph{effective
computational procedure}. A formalization of this observation had
lead famous English mathematician Alan Turing (1936) to the
invention of his celebrated machine and to the corresponding
formalization of the intuitive notion of an \emph{algorithm}. (See
Minsky, 1967, for a relevant discussion of Turing's ideas.)

Now that the concept of an algorithm is defined, we can say that a
model of system (W,D,B), where W is and external memory aid, must
be a \emph{universal computing system}. (This is a necessary but,
of course, not a sufficient, requirement.)\\\\
\textbf{OBSERVATION 2.} We are not born with the knowledge of all
possible algorithms. We can learn, however, to perform, in
principle, any given algorithm, say, by simulating the work of a
Turing machine representing this algorithm.

This observation means that the above system (W,D,B) must
be a \emph{universal learning system}.\\\\
\textbf{OBSERVATION 3.} A person with a good visual memory
performing computations with the use of an external memory aid
learns to perform similar mental computations using the
corresponding imaginary memory aid. A chess player learns to move
chess pieces on an imaginary chess board. An abacus user learns to
operate on an imaginary abacus (Baddeley, 1980). And so on. In
principle, a person can learn to perform any mental computations
by mentally simulating the process of writing symbols on a sheet
of paper.

Ignoring some severe, but theoretically unimportant limitations on
the size of the working space available via this mechanism of
mental imagery, this observation suggests that the human brain, B,
itself -- not just a person with an external memory aid -- must be
treated by a system theorist as a universal learning system.\\\\
\textbf{Note}. An adequate model of B(0) \emph{must} have the
highest general level of computing power. Attempting to simulate
the work of the human brain using a learning system with the
general level of computing power lower than that of the brain can
be compared with an attempt to design a Perpetual Motion machine
in violation of the energy conservation law. No matter how
sophisticated a learning process might be, no system can learn to
do what it cannot do in principle. (An elephant learns to fly only in a Disney film.)\\\\
 \textbf{OBSERVATION 4.}  We (humans) can imagine new sensory events and synthesize
 new motor reactions. At the same time we can remember and recall
 the real sequence of events (reactions). For example,
 an experienced chess player can mentally
 play any chess party. At the same time he/she can recall the real
 parties he/she played. Similarly, we can generate a combinatorial
 number of new  sentences. At the same time we can read by heart
 a specific text we've learned.

 \emph{What kind of learning algorithm can accommodate these
  different types of learning? Do we need different learning algorithms?}\\\\
 \textbf{OBSERVATION 5.} We memorize new information with the
 references to the pieces of the information which we already have in
 our long-term memory (LTM). The more we know in a certain area
 the easier it is to remember new things related to this area.
 For example, we can easily remember long sentences in the language we know. It is
 next to impossible  to remember long sentences in a language we don't know.
 It is also very difficult, for a second language speaker, to get rid of
 the accent, because he/she  tends to build the words of the second
 language from the syllables of the first language.

 \emph{How can this hierarchical referencing system be implemented in neural networks?} \\\\
 \textbf{OBSERVATION 6.} Our ability to retain information in our
 short-term memory (STM) increases if similar information is present in
 our LTM. We can repeat a sentence in the language
 we know. We cannot repeat a sentence in a language we don't know.
 We can imitate  only those reactions of other people that we can do
 ourselves. The same is true for perception. We have difficulties recognizing words
 of a foreign language that we cannot pronounce ourselves.

 \emph{What is STM? How does it interact with LTM? What is working memory?
 What does motor control have to do with it?}\\\\
\textbf{OBSERVATION 7.} To imagine different sensory events we
need to do mental motor reactions that would cause  similar
events. We need to mentally sing a melody to imagine another
person singing this melody. We need to mentally say a sentence to
imagine another person saying this sentence. Etc.

\emph{What is mental imagery? How does mental imagery interact with motor control?}\\\\
\textbf{OBSERVATION 8.} We can see different sub-pictures in the
same picture depending on what we expect to see. The Necker cube
is an example. We can hear different tunes in the same sequence of
sounds (e.g., the sounds produced by a moving train) depending
on what we expect (want) to hear.

\emph{What kind of mechanism available in neural networks can account for these phenomena of mental set?}\\\\
\textbf{OBSERVATION 9.} We can selectively tune our attention to a
voice we want to hear in a noisy room -- the so called \emph{cocktail party phenomenon}.

\emph{ How can the brain temporarily increase sensitivity to signals with some not easily definable characteristics?}\\\\
\textbf{OBSERVATION 10.} Our short-term memory can retain only a
limited number (seven plus or minus two) of items: the "magical
number" of Miller, 1956.  However, due to the effect of "chunking"
the size of a single item can be significantly increased. We also
can  "see more than we can report" ( Sperling, 1960) .
This raises the same set of questions as the Observation 6.\\\\
\textbf{OBSERVATION 11.} The brain is a slow and noisy system. It
cannot process symbolic information in a traditional ("classical")
way by moving symbols in a read/write memory buffer. Nevertheless,
we can learn to mentally simulate different external systems (W,D)
with the properties of a read/write memory. (For example, we can
mentally move chess pieces on an imaginary chess board or mentally
write and erase symbols on an imaginary sheet of paper.)

\emph{How can computational universality in Turing's sense
(Chomsky's  type 0) be achieved without moving symbols in a
read/write memory?} \emph{How can neural networks learn to
simulate a symbolic read/write memory?}\\\\
 \textbf{Note}. The problem of how the brain can \emph{learn to simulate} an external system
(W,D) with the properties of a read/write memory must not be
confused with the problem of how a neural network \emph{can
implement} a read/write memory. The latter problem is trivial. The
former problem is nontrivial and critically important. Traditional
neural network models cannot learn to simulate external systems
with the properties of a read/write memory and, therefore, cannot
serve as models of the
brain's systems responsible for mental imagery.\\\\
\textbf{OBSERVATION 12.} We can recognize that a certain object,
A, is statistically strongly correlated with another object, B. We
can also produce a reaction, R, statistically well correlated with
a certain stimulus, S. Importantly, this statistical relationship
depends on context. Two objects strongly correlated in one context
may be not correlated at all in a different context. Our language
has words \emph{usual, unusual, common, uncommon}, etc., that
reflect our ability to recognize statistical relationships.

\emph{How can a huge amount of computations required for
context-dependent statistical processing be done "on the fly" by
slow neural networks?} (Note that it must be done "on the fly,"
because context can change very rapidly. This statistics cannot be
precalculated, because there is a combinatorial number of possible
contexts!)
\\\\
 \textbf{OBSERVATION 13.} We can wait for a certain object, A.
 Once A appears we recognize that A is the \emph{object we
 were waiting for}. If we expect a certain object, B, to appear and, instead, an
 unexpected object, C, appears we recognize that C is an \emph{unexpected
object}. We can answer the questions: "What are you waiting for?
What do you expect?"

\emph{How does the brain temporarily mark an object as an object
"being waited for" or as an object "being expected?"} \\\\
\textbf{OBSERVATION 14.} Pattern recognition is a
context-dependent activity. Consider the question: "What is it?"
In the context of this question a person behaves as a pattern
classifier. He/she can answer, for example that this is a
\emph{book}. The person's brain was able to distinguish a book
from other objects, say, a \emph{box}, a \emph{disk drive}, etc.
Now consider the instruction: "Take this." In this context it is
no longer important that the object has the name \emph{book}. What
is important is the object's size, weight, position, etc. The
experience acquired while "taking a book" is applicable to "taking
a box" and "taking a disk drive." That is, the same object is
treated as a member of different classes depending on context.

\emph{How can a context-dependent pattern classification be done "on the fly?"}\\\\
\textbf{OBSERVATION 15.} We can recognize our emotional states. We
remember our emotional experience. We use this experience to
evaluate new events. Our concepts of \emph{good, bad, important,
unimportant}, etc. are formed in the process of learning.

\emph{How do we learn to recognize our emotions? How does our
emotional memory interact with other types of memory?}\\\\
\textbf{OBSERVATION 16.} We can recognize internal states and
internal reactions of other people. We can say, for example, "I
know how you feel." We know that another person is
\emph{thinking}, \emph{waiting}, etc. When we learn by imitating
another person, we are not imitating this person as a \emph{black
box}. This means that the problem of learning cannot be formalized
as the automata theory problem of one machine deciphering the
structure of another machine observed as a black box. (If this
formalization were true, we wouldn't be able to learn, in
principle, a behavior of the  Chomsky's type 2 and higher.)

 \emph{How do we learn to control our internal reactions?
 How do we learn the names of our internal reactions
 (thinking, imagining, recalling, waiting, seeing, listening, etc.)?
 How do we recognize  similar internal reactions in other
 people?}\\\\
 \emph{How does mental imagery interact with perception?}\\\\
\textbf{OBSERVATION 17.} Much of what we see we see from our
memory. For example, when we are driving a car in a familiar
environment we need only to glance at the scene to update the
visual picture we expect. We can close our eyes and see the room
we live in by mentally moving the eyes and mentally turning the
head.

\emph{How do the signals coming from external system (W,D)
interact with the signals coming from memory?} \emph{How is our
mental imagery synchronized with the external system (W,D)?}
\emph{What does motor control have to do with it?}

\subsection{Motor control and mental imagery} \label{sec1.4}
\begin{figure}[b!]
\begin{center}
 \includegraphics[width=4.0in]{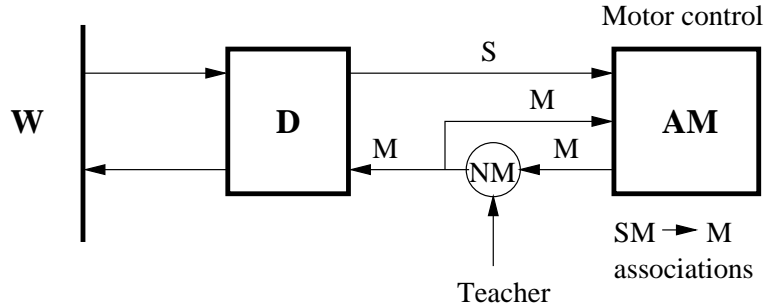}
 \end{center}
 \caption{The concept of forced motor training} \label{fi2}
\end{figure}
Let us expand the structure of system (W,D,B) of Figure \ref{fi1}
as shown in Figure \ref{fi2}. The brain B is divided into two
blocks: AM and NM, where AM is an associative learning system that
forms Sensory,Motor $\rightarrow$ Motor (SM$\rightarrow$M)
associations, and NM is a set of motor centers. The diagram also
depicts the block TEACHER. In this case, the teacher acts as an
idealized neurophysiologists, who can produce any desired output
of centers NM, by "clamping" these centers. System AM receives
sensory signals from system (W,D) and motor signals from the
output of centers NM.  This approach to teaching and learning is
similar to the so-called \emph{supervised learning}, except that,
in our case, the learning system receives its sensory input from
the external system (W,D) rather than from the teacher. This can
be compared with the so-called instrumental conditioning.
\begin{figure}[b!]
\begin{center}
 \includegraphics[width=4.2in]{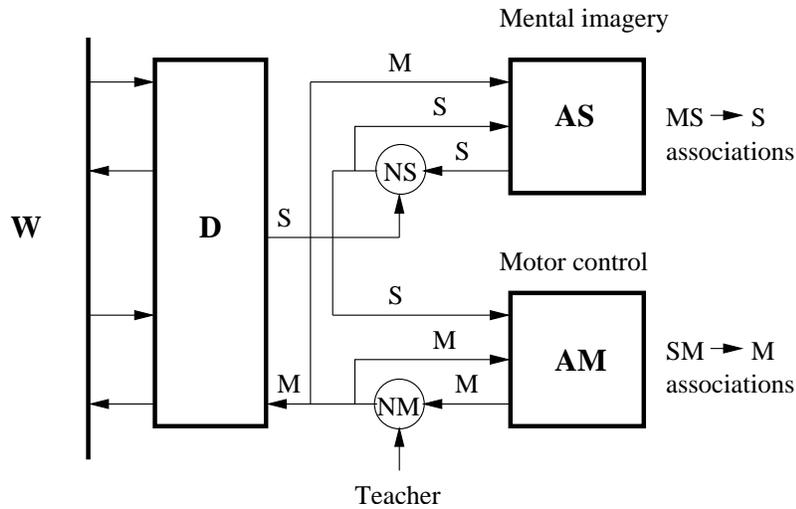}
 \end{center}
 \caption{Mental imagery as a simulation of the external system (W,D)} \label{fi3}
\end{figure}
Let us make the further expansion of the structure of system
(W,D,B) as shown in Figure \ref{fi3}. The brain B is now divided
into four blocks: AS, AM, NS and NM, where blocks AM and NM are
the same as in Figure \ref{fi2}, NS are sensory centers, and AS is
an associative learning system that forms Motor,Sensory
$\rightarrow$ Sensory (MS$\rightarrow$S) associations.

The goal of system AM is to simulate the block TEACHER. The goal
of system AS is to simulate the external system (W,D). It is easy
to see that system (W,D) plays the same role for system AS as the
block TEACHER does for AM. We will view systems AM and AS  as the
systems responsible for motor control and mental imagery,
respectively. We will view the sets of (SM$\rightarrow$M) and
(MS~$\rightarrow$S) associations as the brain's software
associated with the above functions.

\subsection{Mental computations (thinking) as an interaction between motor
control and mental imagery} \label{sec1.5}

\begin{figure}[t!]
\begin{center}
 \includegraphics[width=4.5in]{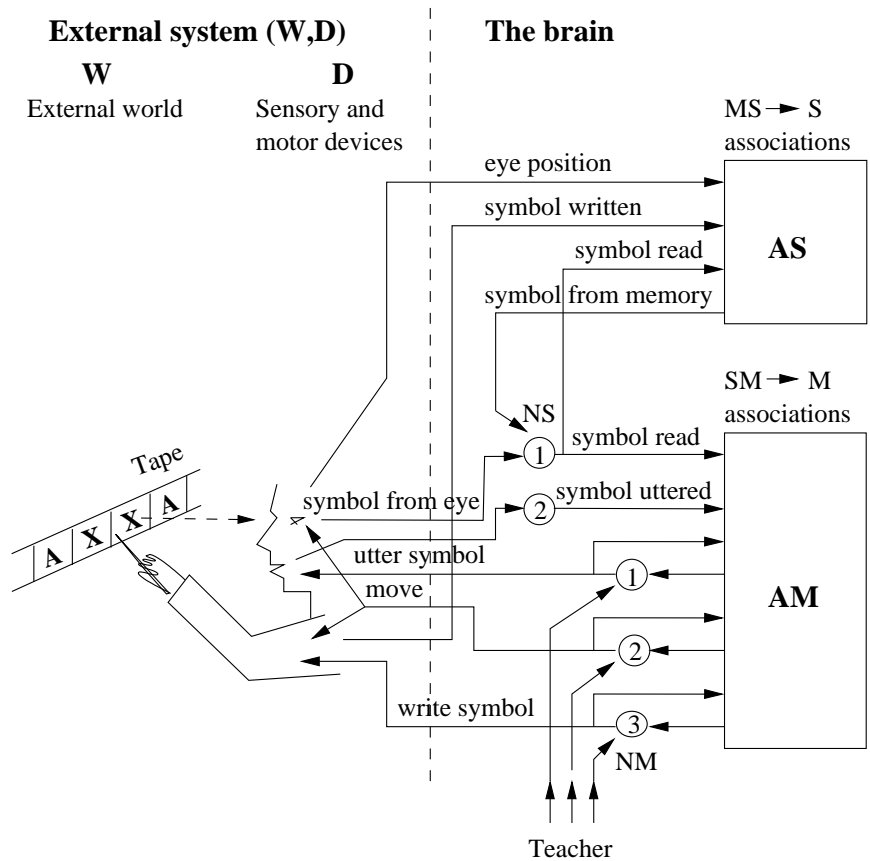}
 \end{center}
 \caption{Mental computations as interaction between motor control and mental imagery} \label{fi4}
\end{figure}

A specific example of system (W,D,B) shown in Figure \ref{fi4}
gives a simplified general explanation of the  phenomenon of
mental computations. The model was implemented as an educational
program, called EROBOT, for the Microsoft Windows. (The program
can be purchased from www.brain0.com.) An explicit description of
this model was given in Eliashberg (2003). In Figure \ref{fi4}
\begin{itemize}
\item W is an external memory aid (the tape divided into squares).
\item D is a set of devices including the eye, the hand and the
speech organ. \item B is the brain divided into four blocks AM,
AS, NM and NS that have the same general meaning as in Figure
\ref{fi3}.
\end{itemize}
The robot's devices, D, allow it to simulate the work of any
Turing machine by performing the following elementary operations:

\begin{enumerate}
\item  read a symbol from the single square scanned by the eye
\item write a symbol into the scanned square \item move the eye
and the hand simultaneously to the next square, the next square
being the one to the left, the one to the right, or the same
square \item utter a symbol to be kept in mind for one cycle --
this one-cycle memory is provided by the delayed feedback between
the motor signal, \emph{utter symbol}, to the speech organ and the
proprioceptive signal,\emph{ symbol uttered}, from this organ.
\end{enumerate}

\noindent An experiment with the model consists of two stages:
training and examination. At the stage of training the teacher
forces the robot (by acting on its motor centers) to perform
several examples of a specified algorithm  with different input
data presented on tape. (The parenthesis checker algorithm
borrowed from Minsky (1967) is used as a built-in example in the
program EROBOT.)

The following results of learning are achieved:
\begin{enumerate}
\item System AM learns to simulate the teacher, so the robot can
perform the demonstrated algorithm with any input data without the
help of the teacher. \item In the case of a finite tape, and a
sufficient number of training examples, system AS learns to
simulate the external system (W,D). Accordingly, the robot learns
to perform the demonstrated algorithm with the use of an imaginary
memory aid. (The robot keeps writing symbols on the real tape to
show what it calculates on the imaginary tape. The robot doesn't
see the real tape!)
\end{enumerate}
\subsection{The pitfall of a "smart" learning algorithm} \label{sec1.6}
 The main part of today's research in learning is devoted to the
 development and study of what can be referred to as "smart" learning
 algorithms. Such algorithms attempt to create "optimal"
 representations of the learner's experience in the learner's memory.
 I argue that this general approach (whatever interesting and important from the
 engineering and mathematical viewpoints) cannot be employed by a
 universal learning system similar to the human brain. The catch is
 that a smart learning algorithm aimed at a "single-context"
 optimization is not universal. While optimizing performance in a
 selected context, it throws away a lot of information needed in a
 variety of other contexts.

Consider, for example, Observation 14 form Section 1.3. This
observation suggests that, in the case of the human brain, there
is no such thing as an optimal context-independent classification.
The main issue is not "how" to pre-process information in the
course of learning (Hebbian learning, backpropagation, simulated
annealing, etc.), and how to store this pre-processed information
in memory (distributed, local, synaptic, optical, etc.), but
"what" information to learn. The human concepts of "good", "bad",
"important", and "unimportant" change with experience. Therefore,
a "smart" learning algorithm with a fixed criterion of optimality
-- the criterion that is not affected by the contents of data --
cannot serve as an adequate metaphor for human learning. What
seems unimportant today may become important tomorrow when new
information is acquired.

I argue that a really smart universal learning system -- such as
B(0) -- must use a "dumb" but universal learning algorithm.
Instead of doing much pre-processing of data before placing it in
memory, such system must use an efficient decision-making (data
interpretation) procedure to process "raw experience" dynamically
(on the fly) depending on context. Theoretically, a powerful
enough interpretation procedure can always make up for a "dumb"
learning algorithm as long as this algorithm doesn't lose data. In
contrast, no decision making procedure can make up for a "smart"
learning algorithm that throws away a lot of information. The loss
of data is irremediable.

\section{From Associative Neural Networks to E-machines} \label{sec2}

This section introduces the concept of a primitive E-machine
(Eliashberg, 1979) as a natural information processing extension
of the notion of a homogeneous associative neural network. A
complex E-machine is a system built from several primitive
E-machines. Complex E-machines will be discussed in Part II of
this paper.

\subsection{ Simple example of associative neural network: Model ANN-0} \label{sec2.1}
\begin{figure}[b!]
\begin{center}
 \includegraphics[width=4.0in]{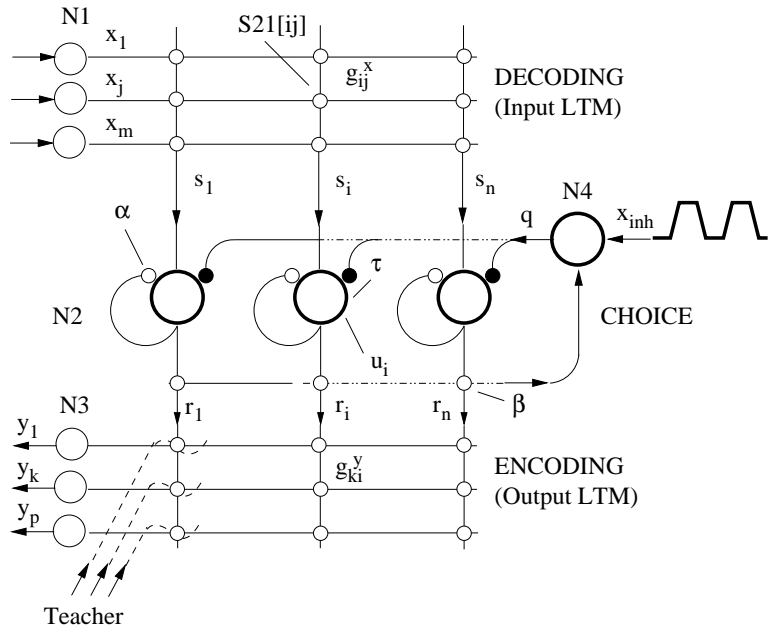}
 \end{center}
 \caption{Simple example of associative neural network} \label{fi5}
\end{figure}

Consider a neural network schematically shown in Figure \ref{fi5}.
The functional model of this network described in this section
will be referred to as Model ANN-0 (Associative Neural Network \#
0).

In Figure \ref{fi5}, large circles with incoming and outgoing
lines represent neurons with their dendrites and axons,
respectively. Small white and black circles represent excitatory
and inhibitory synapses, respectively. The network has three
layers of neurons: input neurons N1, intermediate neurons N2, and
output neurons N3. Neurons N2 have a global inhibitory feedback
via neuron N4 and local excitatory feedbacks. It will be shown
that in this network neurons N2 can compete via reciprocal
inhibition in the \emph{winner--take--all} fashion. A similar
effect can be obtained in a network with lateral inhibitory
feedbacks. Figure \ref{fi5} uses the following notation:
\setcounter{equation}{0}

\begin{itemize}
\item $Nk[j]$ is the $j$-th neuron from set $Nk$. \item $Smk[i,j]$
is the synapse between neuron $Nk[j]$ and neuron $Nm[i]$.\item
$x_{j}$ is the output of neuron $N1[j]$. \item $g_{ij}^{x}$ is the
gain of synapse $S21[ij]$. \item $s_{i}$ is the net synaptic
current of synapses $S21[i,1],\ldots S21[i,m]$ -- $s_{i}$
represents a similarity between input vector $x$ and vector
$g_{i}^{x}$ (expression (\ref{eq2.1})). \item $r_{i}$ is the
output of neuron $N2[i]$. \item $q$ is the output of neuron N4.
This output is the sum of the feedback signal $\beta\sum r_{i}$
and an external signal $x_{inh}$. \item $\beta$ is the gain of
synapse between any neuron from N2 and neuron N4. \item $\tau$ is
the time constant of any neuron from N2. \item $\alpha$ is the
gain of synapse providing local excitatory feedback for a neuron
from N2. \item $g_{ki}^{y}$ is the gain of synapse between neuron
$N2[i]$ and neuron $N3[k]$.
\end{itemize}
The following functional model of the network of Figure \ref{fi5}
was studied in Eliashberg (1967, 1979). In this model a neuron is
treated as a linear threshold element with zero threshold and the
time constant $\tau$. In spite of its simplicity, this model has a
significant educational value because it allows one to explicitly
bridge the gap between its neurobiological and psychological
theories and to show what kind of mathematics is involved in this
bridging. No learning algorithm is described, and it is assumed
that the model is preprogrammed before the beginning of an
experiment.

 \setcounter{equation}{0}
\begin{equation} \label{eq2.1}
  s_i = \sum_{j=1}^{m}{g_{ij}^{x}\cdot x_{j}}
\end{equation}
\begin{equation} \label{eq2.2}
  \tau \frac{du_{i}}{dt}+u_{i}= s_{i}+\alpha \cdot r_{i}-q
\end{equation}
\begin{equation} \label{eq2.3}
   r_{i}=
   \begin{cases}
     u_{i} \quad &\text{if $u_{i}>0$} \\
     0 \quad &\text{otherwise} \\
   \end{cases}
\end{equation}
\begin{equation} \label{eq2.4}
  q=\beta \sum_{i=i}^{n}{r_{i}+x_{inh}}
\end{equation}
\begin{equation} \label{eq2.5}
  y_k = \sum_{i=1}^{n}{g_{ki}^{y}\cdot r_{i}}
\end{equation}
Let all $x_{j}$ and $x_{inh}$ (and, therefore, all $s_{i}$) be
step functions of time. Then, for all active neurons from layer N2
-- the neurons for which $u_{i}>0$ -- the solution of equations
(\ref{eq2.2})--(\ref{eq2.4}) can be represented in the following
explicit form:
\begin{gather}\label{eq2.6}
\begin{split}
&u_{i}=\frac{(s_{i}-s_{av})}{\alpha-1}(e^{at}-1)+(u_{i}^{0}-u_{av}^{0})e^{at}
\\ &+\frac{(s_{av}-x_{inh})}{1+\beta \cdot n_{1}-\alpha}(1-e^{-bt})+u_{av}^{0}\cdot e^{-bt}
\end{split}
\end{gather}
where
\begin{itemize}
\item $n_{1}$ is the number of active neurons from N2. \item
$u_{i}^{0}$ $(i=1,\ldots n)$ are the values of $u_{i}$ at t=0.
\item $s_{av}$ and $u_{av}^{0}$ are the average values of $s_{i}$
and $u_{i}^{0}$ for all active neurons from N2.
\end{itemize}
\begin{equation} \label{eq2.7}
  s_{av}= \frac{1}{n_{1}} \sum_{i=1}^{n_{1}}s_{i}
\end{equation}
\begin{equation} \label{eq2.8}
  u_{av}^{0}= \frac{1}{n_{1}} \sum_{i=1}^{n_{1}}u_{i}^{0}
\end{equation}
Parameters $a$ and $b$ in $e^{at}$ and $e^{-bt}$ are as follows:
\begin{equation} \label{eq2.9}
   a = (\alpha -1)/\tau
\end{equation}
\begin{equation} \label{eq2.10}
  b=(1+\beta \cdot n_{1} -\alpha )/\tau
\end{equation}
Let $1<\alpha <1+\beta$. Then $a>0$. According to expression
(\ref{eq2.6}), neurons $N2[i]$ with $s_{i}>s_{av}$ increase their
potentials $u_{i}$. Neurons $N2[i]$ with $s_{i}<s_{av}$ decrease
their potentials and switch off once $u_{i}<0$. This reduces
$n_{1}$ and increases $s_{av}$ making $s_{i}<s_{av}$ for some
additional neurons from N2. Eventually, only neurons with
$s_{i}=max(s_{1},\ldots s_{n})$ will have $u_{i}>0$. It can be
shown that this equilibrium is unstable if $n_{i}>1$. Therefore,
in the presence of noise, at the end of the transient response
there will be only one winner randomly selected from the set of
neurons with the maximum level of $s_{i}$.

\subsection{Model ANN-0 as a symbolic machine} \label{sec2.2}
 Let us introduce a finite ("psychological") time step $\Delta t \gg \tau$, and let us assume
 that inputs change step-wise at  moments $t_{\nu}$ and $t_{\nu}+\Delta t /2 $, where
\begin{equation} \label{eq2.11}
  t_{\nu}=\nu \cdot \Delta t \qquad  \nu = 0,1,\ldots
\end{equation}

\begin{equation} \label{eq2.12}
   x_{j}(t)=
   \begin{cases}
   \overline{x}_{j}(\nu) \quad  &\text{for $ t \in (t_{\nu},t_{\nu}+\Delta t /2] $}\\
   0 \quad &\text{otherwise}
   \end{cases}
\end{equation}
 Let us introduce a periodic inhibition
\begin{equation}
   x_{inh}=
   \begin{cases}
     x_{inh}^{0}  \quad &\text{if $ t \in (t_{\nu},t_{\nu} + \Delta t /2] $} \\
     0 \quad &\text{otherwise} \label{eq2.13}\\
\end{cases}
\end{equation}
Let us sample outputs at the end of the first half of each cycle
\begin{equation} \label{eq2.14}
  \overline{y}_{k}(\nu) = y_{k}(t_\nu+\Delta t/2)
\end{equation}
Let us assume that the the states of ILTM and OLTM are specified
at the beginning of an experiment with the model and don't change
during the experiment (the model is preprogrammed in advance and
no learning takes place during the experiment). Let us also assume
that the parameters of the model are the same for all experiments.
To describe the "psychological" properties of Model ANN-0 we need
the following system theoretical concepts.\\\\
\textbf{DEFINITIONS}:
\begin{itemize}
\item A (deterministic) \emph{combinatorial machine} is a system
M=(\textbf{X},\textbf{Y},f), where \textbf{X} and \textbf{Y} are
finite sets of symbols, called the \emph{input} and the
\emph{output} set (or \emph{alphabet}) of M, respectively;
$f:\textbf{X}\rightarrow \textbf{Y}$ is the \emph{output function}
of M. Machine M works as follows: $ y_{\nu}= f(x_{\nu})$, where
$x_{\nu} \in \textbf{X}$ and $y_{\nu} \in \textbf{Y}$ are the
input and the output symbols at the $\nu$-th cycle.

\item A \emph{probabilistic combinatorial machine} is a system
M=(\textbf{X},\textbf{Y},$\delta$), where \textbf{X} and
\textbf{Y} are the same as above; $\delta :\textbf{X} \times
\textbf{Y} \rightarrow [0,1]$ is the \emph{function of output
conditional probabilities} of M. Machine M works as follows:\\ $
P\{y_{\nu}=b \ |\ x_{\nu}=a \} = \delta (a,b)$, where $x_{\nu},a
\in \textbf{X}$ and $y_{\nu},b \in \textbf{Y}$ and $P \{B\ |\ A
\}$ is the conditional probability of B given A. \item Machine M1
\emph{simulates} (\emph{is equivalent to}) machine M2 if these two
machines cannot be distinguished from each other by observing
their inputs and outputs (observing them as black boxes).
\end{itemize}
\bigskip
\noindent The following properties of Model ANN-0 -- with the
inputs and outputs described by expressions \ref{eq2.11},
\ref{eq2.12}, \ref{eq2.13}, \ref{eq2.14} -- can be proved
(Eliashberg, 1979):
\begin{enumerate}
\item Let \textbf{X} and \textbf{Y} be finite subsets of the sets
of input an output vectors of the model, respectively. Let
$\bar{x}(\nu) \in \textbf{X}$ and $\bar{y}(\nu) \in \textbf{Y}$.
Let $f^{s}: \textbf{X}\times \textbf{X}\rightarrow \textbf{R}$ be
the similarity function from expression (\ref{eq2.1})-- in this
case $f^{s}$ is the scalar product. Let the pair
$(\textbf{X},f^{s})$ satisfy the following \emph{correct decoding
condition}
\begin{equation} \label{eq2.15}
  \forall x,x' \in \textbf{X} \qquad (\text{if $x \neq x'$ then $f^s(x,x') <
  f^s(x,x)$)}
\end{equation}
 For any combinatorial machine
$M=(\textbf{X},\textbf{Y},f)$ there exists a state, $g$, of the
LTM of the model (and some fixed values of parameters of the
model) such that the model in the state $g$ simulates (is
equivalent to) machine M. \item The previous result extends to any
probabilistic combinatorial machine
(\textbf{X},\textbf{Y},$\delta$) with rational probabilities
$\delta$.
\end{enumerate}
\subsection{Model AF-0: A trivial primitive E-machine corresponding to Model ANN-0} \label{sec2.3}

The "psychological" properties of Model ANN-0 can be described in
algorithmic terms. The description presented below gives an
example of a \emph{trivial primitive E-machine} -- a primitive
E-machine without E-states. This model will be referred to as
Model AF-0 (Associative Field \# 0).\\\\
\textbf{Notation}\\
In this paper I use a C-like notation mixed with scientific-like
notation to represent models of E-machines aimed at humans. (I use
C++ for computer simulation.) I use special notation for the
following operations:
\begin{itemize}
\item $ A:=\{a | P(a)\}$ \quad  select the set of elements $a$
with the property $P(a)$. I use Pascal-like notation ":=" to
emphasize the dynamic character of this operation. \item $a: \in
A$ \quad select an element $a$ from the set $A$ at random with
equal probability.
\end{itemize}

\bigskip

\noindent \textbf{DECODING}: \quad compare input vector with all vectors in Input LTM\\\\
   $for(i=1;i<=n;i++) \quad s[i]=Similarity(x[*],gx[*][i]);$   \qquad (1)\\\\
\textbf{CHOICE}:\quad  select the set of locations with the maximum value of $s[i]$\\\\
   $ MAXSET:=\{i \ |\ s[i]=max(s[1], \ldots s[n]) \}; $ \qquad (2)\\\\
randomly select a winner (win) from MAXSET\\\\
   $ win : \in  MAXSET; $ \qquad(3)\\\\
\textbf{ENCODING}: \quad read output vector from the selected location, $win$, of Output LTM\\\\
 $if(s[win] > xinh) \quad y[*] = gy[*][win]; \quad else \quad  y[*]=NULL;$ \qquad (4)\\\\
\textbf{Comments}:
\begin{enumerate}
\item  As long as the $Similarity()$ function and the set of
allowable inputs, \textbf{X}, satisfy the correct decoding
condition (expression (\ref{eq2.15}) with $f^s= Similarity $),
Model AF-0 is a system universal with respect to the class of
combinatorial machines. \item The psychological Model AF-0 is much
simpler than the neurobiological model ANN-0. Model AF-0 doesn't
have all the \emph{neural-implementation-parameters} of model
ANN-0. It also doesn't have the fast changing (neurobiological)
state $u$. \item In the next sections Model AF-0 will be enhanced
in several directions.
    \begin{enumerate}
    \item Adding a \emph{one-cycle delayed feedback} from $y$ to
    $x$. This will change model AF-0 into a system universal with
    respect to the class of state machines.
    \item Adding a \emph{universal learning algorithm}. The new
    model will become a learning system universal with respect to
    the class of finite-state machines.
    \item Introducing  \emph{E-state arrays}, a \emph{next E-state
    procedure}, and a \emph{Structural LTM (SLTM)}. This will transform
    Model AF-0 into a nontrivial primitive E-machine capable of
    producing some interesting effects of working memory and temporal context
    (mental set).
    \item  Introducing \emph{associative inputs and outputs}. This
    enhancement will allow us to get effects of sparse re-coding, data
    compression and context-dependent statistical filtering.
    \end{enumerate}
\end{enumerate}

\subsection{Delayed feedback and simulation of finite-state machines} \label{sec2.4}

\textbf{DEFINITION}\\  A (deterministic) finite--state machine is
a system $M=(\textbf{X},\textbf{Y},\textbf{S},\alpha,\omega)$,
where \textbf{X} and \textbf{Y} are finite sets of external
symbols of M called the \emph{input and the output sets
(alphabets)}, respectively, \textbf{S }is a finite set of internal
symbols of M called the \emph{state set}, $\omega:\textbf{X}
\times  \textbf{S} \rightarrow \textbf{Y}$ is a function called
the \emph{output function} of M, $\alpha:\textbf{X} \times
\textbf{S} \rightarrow \textbf{S}$ is a function called the
\emph{next-state function} of M. The work of machine M is
described by the following expressions:
$s_{\nu+1}=\alpha(x_{\nu},s_{\nu})$, and
$y_{\nu}=\omega(x_{\nu},s_{\nu})$, where $x\in\textbf{X}$,
$y\in\textbf{Y}$, and $ s \in \textbf{S}$ are the values of input,
output, and state variables at the moment $\nu $, respectively.

\textbf{Note}. There are different equivalent formalizations of
the concept of a finite--state machine. The formalization
described above is known as a Mealy machine. Another popular
formalization is a Moore machine. In a Moore machine the output is
described as a function of the next--state. Practical electronic
designers usually use the term \emph{state machine} instead of the
term \emph{finite--state machine}.
\begin{figure}[t!]
\begin{center}
 \includegraphics[width=2.5in]{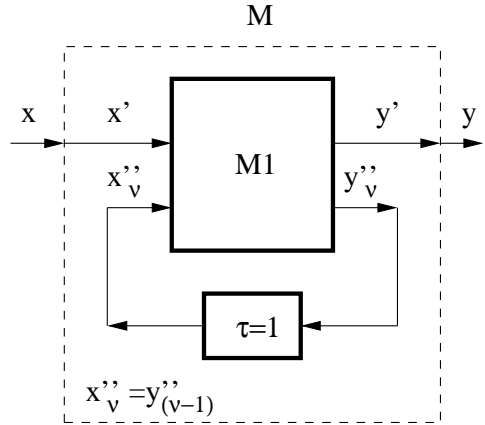}
 \end{center}
 \caption{Finite--state machine as a combinatorial machine with a one-cycle delayed feedback} \label{fi6}
\end{figure}
Any finite--state machine can be implemented as a combinatorial
machine with a one cycle delayed feedback (see Figure \ref{fi6}).
Using this trick, it is easy to show that Model AF-0 with a
delayed feedback can simulate any finite--state machine.

\subsection{Introducing a universal learning algorithm} \label{sec2.5}

Let us return to the system (W,D,B) shown in Figure \ref{fi4}.
Simple as it is, Model AF-0 has enough computing power to serve as
the motor control system AM, because the one-cycle delayed
feedback "utter-symbol$ \rightarrow $ symbol-uttered" transforms
block AM into a system universal with respect to the class of
finite machines (as explained in the previous section). This gives
the system (W,D,B) the power of a universal Turing machine. (A
Turing machine is a finite--state machine coupled with an external
\emph{tape} through the I/O device called the \emph{head}.  The
block (W,D) provides the functionality of the tape and the head of this machine.)\\\\
\emph{What kind of learning algorithm does the Model AF-0 need to
be able to learn to simulate any combinatorial machine?}\\\\
It is easy to show that the simplest algorithm satisfying this
requirement is \emph{"tape-recording"} the X-~sequence and the
Y-sequence in the Input and Output LTM, respectively. In the case
of a deterministic combinatorial machine, this algorithm can be
improved by recording only new associations. In the case of a
probabilistic combinatorial machine the same associations need to
be recorded several times to accumulate statistics.

In phenomenological terms, the  above tape recording algorithm can
be described as follows:\\\\
$BOOL  \quad wen;$ \qquad //write enable: auxiliary input variable \\
$int  \quad wptr;$ \qquad  \quad // write pointer: auxiliary state variable\\\\
  $ if(wen) \quad \{ gx[*][wptr]=x[\ast]; \quad  gy[*][wptr]=y[*];  \quad  wptr++;
  \}$ \qquad (5)\\\\

It is interesting to mention that some famous psychiatrists were
advocating this concept of \emph{tape-recording-learning}. Here is
a quotation from Meynert (1884): "Each new impression meets a new,
still vacant cell. With the existence of such vast number of these
vacant cells, impressions arriving in succession find carriers in
which they will remain forever in the same close order".

 As mentioned in \ref{sec1.6}, the concept of a "smart" learning algorithm
 creates a methodological pitfall. The catch is that the human concept of
 \emph{important information} changes with  experience, so no learning
 algorithm with a fixed criterion of optimality can be smart enough to
 know in advance which information is important to store and which is not.
 What seems unimportant today may become very important tomorrow
 where more information is acquired.

 I argue that there is no special magic in \emph{how} the
 knowledge is stored in the brain (distributed, local, analog, digital,
 etc.). The magic is in \emph{what} knowledge is stored and how this
 knowledge is processed dynamically depending on context.

\subsection{"Symbolic" or "nonsymbolic," that is the question} \label{sec2.7}
Starting with the neural network shown in Figure \ref{fi5}, one
can proceed  in two different directions:
\begin{enumerate}

\item When the neurons in layer N2 compete in a
\emph{winner-take-all} fashion ($1<\alpha<1+\beta$), the Model
ANN-0 can be thought of as a neural counterpart of the
Programmable Logic Array (PLA) shown in Figure \ref{fi7}.
\begin{figure}[t!]
\begin{center}
 \includegraphics[width=4.0in]{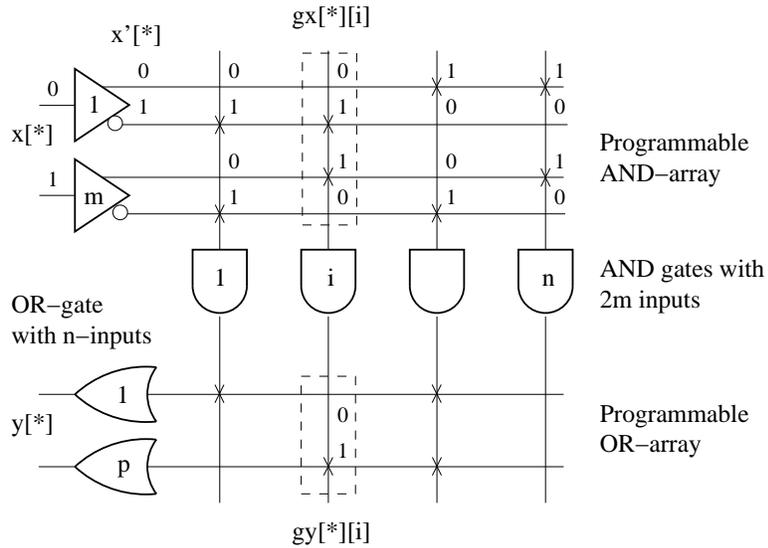}
 \end{center}
 \caption{Programmable Logic Array (PLA)} \label{fi7}
\end{figure}
The input synaptic matrix (Input LTM) is similar to the
programmable AND--array, and the output synaptic matrix (Output
LTM) is similar to the programmable OR--array. If one goes in this
direction one  gets some "neural extras," such as a generalization
by similarity and the ability to simulate probabilistic
combinatorial machines.
\begin{figure}[b!]
\begin{center}
 \includegraphics[width=2.5in]{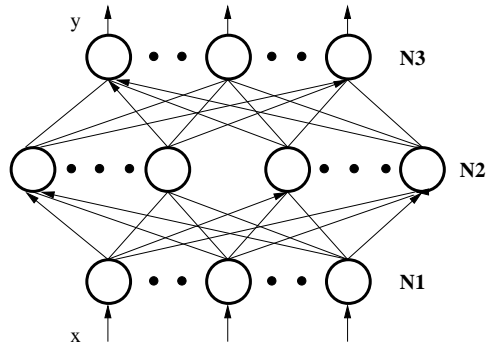}
 \end{center}
 \caption{Associative neural network as a "connectionist" system} \label{fi8}
\end{figure}
Taking this path, eventually, brings one to the concept of a
primitive E-machine. \item If one reduces the competition of
neurons in layer N2, one enters the realm of \emph{connectionist}
neural networks. Let us set $\alpha=\beta=0$. Let us also replace
the linear threshold output function by a sigmoid function. Model
ANN-0 becomes a typical Parallel Distributed Processing (PDP)
system. In the traditional connectionist graphical representation,
this system looks like shown in Figure~\ref{fi8}.

If one selects this "nonsymbolic" path, one is inspired to view
neural networks as analog computational devices implementing
multidimensional mappings $ f: \textbf{R}^m \rightarrow
\textbf{R}^n $, where $\textbf{R}$ is the set of real numbers. It
is seldom possible to find the weights, corresponding to
nontrivial multidimensional mappings, analytically. Therefore the
development and study of the  learning algorithms, automatically
adjusting the weights, becomes the main thrust of this research.
(There is plenty of room in multidimensional real spaces, so one
can spent one's life searching for the "magical neural mappings.")
\end{enumerate}

\noindent \emph{Which way to go?} I argue that the first direction
is the right way to go if one is interested in biological brain.
The second approach has the following liabilities (each of which
is sufficient to disqualify this approach as an adequate
biological framework):
\begin{enumerate}
\item The learning algorithms used in PDP models (such as
backpropagation, simulated annealing, etc.) are not universal.
(See Rumelhart, McClelland, et al (1986) for the explanation of
the PDP framework.)
 \item Traditional PDP models don't have a
sufficient general level of computing power to adequately address
such critically important "symbolic" problems as the problem of
\emph{natural language}. (See Pinker and Mehler  (1988) for a
discussion of this issue.) \item
 PDP models provide no satisfactory explanation of the phenomena
 of working memory and mental set. They are largely inconsistent with
 Observations 1-17 from Section 1.5.
\item Biological neural networks don't have the accuracy needed to
implement traditional PDP algorithms. \item Traditional PDP models
have no room to accommodate the known complexity of biological
neurons. The whole vision of the brain as a
\emph{collective-distributed-dynamical} system built from simple
"atomic" neurons is inconsistent with the modern neurobiological
data (Kandel and Spenser, 1968; Kandel, Jessel, and Schwartz,
2000; Nichols, Martin, Wallace, 1992, Byrne, 1987).  A single
neuron is a complex integrated computing element. The brain has
many different types of neurons tailored for different tasks.
\end{enumerate}

\subsection{Introducing E-states: Model AF-1} \label{sec2.8}

\begin{figure}[b!]
\begin{center}
 \includegraphics[width=5.5in]{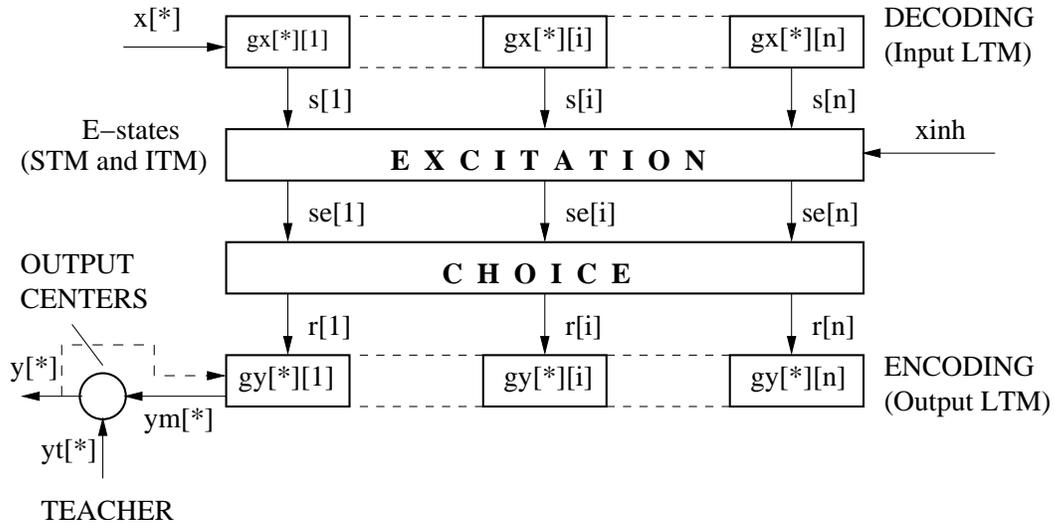}
 \end{center}
 \caption{The general architecture of Model AF-1} \label{fi9}
\end{figure}

The basic architecture of Model AF-1 is shown in Figure \ref{fi9}.
As compared with Model AF-0, this model has two additional
procedures: BIAS and NEXT E--STATE PROCEDURE. Both these
procedures are included in the block EXCITATION.\\\\
\textbf{DECODING}: \quad compare input vector with all vectors in Input LTM\\\\
   $for(i=1;i<=n;i++) \quad s[i]=Similarity(x[*],gx[*][i]);$   \qquad (1)\\\\
\textbf{BIAS}: \quad calculate biased similarity. Coefficients $a$
and $b$ determine, respectively, \\the \emph{additive} and the
\emph{multiplicative} biassing effect of the "residual excitation" $e[i]$.\\\\
   $for(i=1;i<=n;i++) \quad se[i]= s[i] + a*e[i] + b*s[i]*e[i]);$   \qquad (2)\\\\
\textbf{CHOICE}:\quad  select the set of locations with the maximum value of
$se[i]$\\\\
   $ MAXSET:=\{i \ |\ se[i]=max(se[1], \ldots se[n]) \}; $ \qquad (3)\\\\
randomly select a winner (win) from MAXSET\\\\
   $ win : \in  MAXSET; $ \qquad(4)\\\\
\textbf{ENCODING}: \quad read output vector from the selected location, $win$, of Output LTM\\\\
 $if(s[win] > xinh) \quad y[*] = gy[*][win]; \quad else \quad  y[*]=NULL;$ \qquad (5)\\\\
\textbf{NEXT E-STATE PROCEDURE}: \quad calculate next E-state\\\\
$for(i=1;i<=n;i++)$ \ \qquad  \qquad \qquad \qquad \qquad (6) \\$
if (s[i]> e[i])\quad  e[i]= s[i];$ \qquad \qquad //instant charge
\\ $ else \quad e[i]= e[i]*(tau-1)/tau ;$   \qquad //discharge
with the time constant $tau$\\\\
\textbf{LEARNING}: \quad calculate next state of LTM (G-state)\\\\
 $ if(wen) \quad \{ gx[*][wptr]=x[\ast]; \\ \quad  gy[*][wptr]=y[*];  \quad  wptr++;
  \}$ \qquad (7)\\\\

For the sake of concreteness let us define the following
similarity function:\\\\
$float \quad Similarity(int *x,int *g)$ \qquad \qquad (8)\\
$\{ $\\
\indent $float \quad s;$\\
\indent $int \quad j,k;$ \\
\indent $ s=0; \quad k=0;$\\
\indent $for(j=1;j<=m;j++)$\\
\indent $\{if(x[j]==g[j]\quad \& \& \quad x[j]\ !=\ 0) \quad s++;$\\
\indent $if(x[j]\ != \ 0) \quad k++;\}$\\
\indent $if(k>0) \quad s \ /=\ k; \quad else \quad s=0;$\\
\indent $return \quad s;$\\
$\}$ \\
\textbf{Note}. The Similarity() is equal to the number of non-zero
matches ($x[j]=g[j]\neq~0$) divided by the number of non-zero
components of input vector ($x[j]\neq 0$). Many other similarity
functions, satisfying the correct decoding condition -- Section
2.2, Expression (\ref{eq2.15}) -- would work as well.
\subsection{Dynamic reconfiguration: "many symbolic machines in one"} \label{sec2.9}
Model AF-1 uses a very simple mechanism of EXCITATION (expressions
(2) and (6)). This simple mechanism is sufficient to illustrate
some important effect associated with the introduction of
E-states.\\\\
\textbf{Terminology}. The pair (gx[*][*],gy[*][*]) will be called
the \emph{program} or the \emph{table of associations} of Model
AF-1. The pair (gx[*][i],gy[*][i]) will be called the $i-th$
\emph{command} (the $i-th$ \emph{association}) of the program (the
table of associations). The number of commands in a program will
be called the \emph{length of the program}.\\\\
It is easy to prove the following result:

 Let C(\textbf{X},\textbf{Y}) be a class of combinatorial
machines with the input alphabet \textbf{X} and the output
alphabet \textbf{Y}. Model AF-1 with a fixed program of the length
$|\textbf{X}|  \cdot  |\textbf{Y}| $ (or greater) can be changed
(reconfigured) into any machine from  class
C(\textbf{X},\textbf{Y}) by changing its E-state, $e[\ast]$.

\textbf{Proof}. Let the program $(gx[\ast][\ast],gy[\ast][\ast])$
contain at least once each pair from \textbf{X}$\times$\textbf{Y},
and let \textbf{N}(M) be the subset of locations containing all
commands of a combinatorial machine M from the above class. Let
$\tau \gg 1$. Let $e[i]=1$ if $i\in \textbf{N}(M)$, and $e[i]=0$
otherwise. Model AF-1 with this program and this E-state will
simulate machine M.

This result illustrates the importance of E-states. Model AF-0
with a program of the length $|\textbf{X}|$ can simulate a single
combinatorial machine from C(\textbf{X},\textbf{Y}). To simulate a
different machine, this model must be reprogrammed. Model AF-1
with a program of the length $|\textbf{X}|  \cdot  |\textbf{Y}| $
can simulate any machine from the above class without
reprogramming.\\\\
\textbf{Example}.  Let C(\textbf{X},\textbf{Y}) be the class of
all logic functions with  $m$ inputs and one output, that is,
$\textbf{X}=\{0,1\}^m$ and  $\textbf{Y}=\{0,1\}$. Model AF-1 with
a fixed program of the length $2m$ can be reconfigured into any of
the $2^N$ possible logic functions, where $N=2^m$.

\emph{Why is it better to reconfigure than to reprogram?} This
critically important question will be discussed in  Part II of
this paper.
\section{Molecular Interpretation of E-states: Ensembles of
Protein Nanomachines as Statistical Mixed-signal Computers}
\label{sec3} \emph{What can be a meaningful neurobiological
interpretation of the phenomenological E-states? How can
nontrivial next E-state procedures be implemented in neural
networks?}

This section presents a formalism that offers an answer to these
questions. The formalism can be viewed as a system theoretical
extrapolation of the main idea of the Hodgkin and Huxley (1952)
theory that the sodium and potassium ion channels, embedded in the
axon membrane, work as stochastic switches with several internal
states (conformations). The formalism was discussed in Eliashberg
(1989, 1990a, and 2003).

\subsection{ Concept of protein molecule machine (PMM) }
\label{sec3.1}
\begin{figure}[b!]
\begin{center}
\includegraphics[width=4.0in]{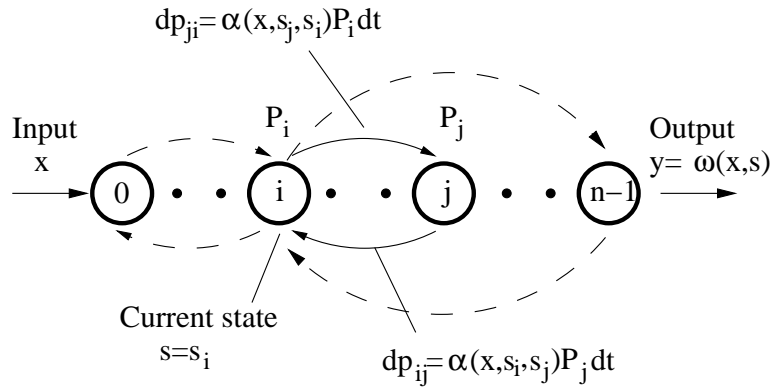}
\end{center}
\caption{Protein molecule as a probabilistic nanomachine}
\label{fi10}
\end{figure}
 \textbf{DEFINITION}. A \emph{Protein Molecule Machine} (PMM) is
 an abstract probabilistic computing system
$(\textbf{X},\textbf{Y},\textbf{S},\alpha,\omega)$, where
\begin{itemize}
\item \textbf{X} and \textbf{Y} are the sets of real input and
output vectors, respectively  \item \textbf{S}$=\{
s_{0},..s_{n-1}\}$ is a finite set of states  \item $\alpha:
\textbf{X}\times\textbf{S}\times\textbf{S} \to \textbf{R}^\prime$
is a function describing the input-dependent conditional
probability densities of state transitions, where
$\alpha(x,s_i,s_j)dt$ is the conditional probability of transfer
from state $s_j$ to state $s_i$ during time interval $dt$, where
$x\in \textbf{X}$ is the value of input, and $\textbf{R}^\prime$
is the set of non-negative real numbers. The components of $x$ are
called \emph{generalized potentials}. They can be interpreted as
membrane potential, and concentrations of different
neurotransmitters.

\item $\omega: \textbf{X}\times\textbf{S}\rightarrow\textbf{Y}$ is
a function describing output. The components of $y \in \textbf{Y}$
are called \emph{generalized currents}. They can be interpreted as
ion currents, and the flows of second messengers.
\end{itemize}
\noindent Let $x\in\textbf{X}$, $y\in\textbf{Y}$, $s\in\textbf{S}$
be, respectively, the values of input, output, and state at time
t, and  let $P_{i}$ be the probability that $s=s_i$. The work of a
PMM is described as follows:
\setcounter{equation}{0}
\begin{align}
   &\frac{dP_i}{dt} = \sum_{j\neq i}{\alpha(x,s_i,s_j)P_j } - P_i \sum_{j\neq i}{\alpha(x,s_j,s_i)}
   \label{eq3.1}\\
   &at\  t=0 \qquad \sum_{i=0}^{n-1}{P_i} = 1 \label{eq3.2}\\
   &y =\omega(x,s) \label{eq3.3}
\end{align}

\noindent Summing the right and the left parts of (\ref{eq3.1})
over $i=~0,..n-1$ yields
\begin{equation} \label{eq3.4}
    \frac{d (\sum_{i=0}^{n-1}{P_i })}{dt}= 0
\end{equation}

\noindent so the condition (\ref{eq3.2}) holds for any t.

\noindent The internal structure of a PMM is shown in Figure
\ref{fi10}, where $dp_{ij}$ is the probability of transition from
state $s_{j}$ to state $s_{i}$ during time interval $dt$. The
output $y=~\omega(x,s)$ is a function of input and the current
state.

\noindent For the probability of transition from state $s_j$ to
state $s_i$ we have
\begin{equation} \label{eq3.5}
   dp_{ij} = \alpha(x,s_i,s_j) P_j dt
\end{equation}
\noindent It follows from (\ref{eq3.1}) that
\begin{equation} \label{eq3.6}
  dP_i = \sum_{j\neq i}{(dp_{ij}-dp_{ji})}
\end{equation}
\subsection{Example: Voltage-Gated Ion Channel as a PMM}
\label{sec3.2} Ion channels are studied by many different
disciplines: biophysics, protein chemistry, molecular genetics,
cell biology and others (see Hille, 2001). I am concerned with the
information processing (computational) possibilities of ion
channels.

I postulate that, at the information processing level, ion
channels (as well as some other membrane proteins) can be treated
as PMMs. That is, at this level, the exact biophysical and
biochemical mechanisms are not important. What is important are
the properties of ion channels as abstract machines.

This situation can be meaningfully compared with the general
relationship between statistical physics and thermodynamics. Only
some properties of molecules of a gas (e.g., the number of degrees
of freedom) are important at the level of thermodynamics.
Similarly, only some properties of protein molecules are important
at the level of statistical computations implemented by the
ensembles of such molecules.

\medskip

\begin{figure}[t!]
\begin{center}
\includegraphics[width=5.0in]{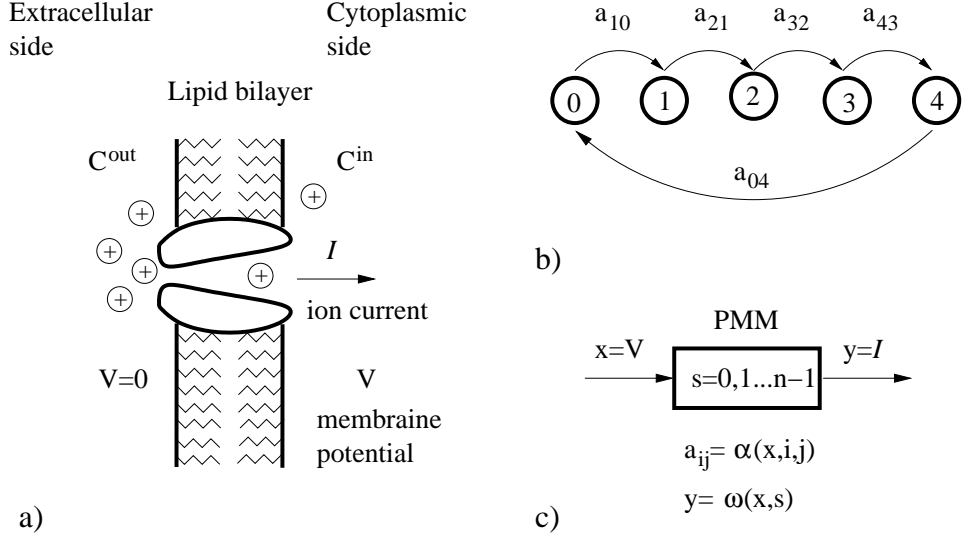}
\end{center}
\caption{Ion channel as a PMM} \label{fi11}
\end{figure}

The general structure of a voltage-gated ion channel is shown
schematically in Figure \ref{fi11}a. Figures \ref{fi11}b and
\ref{fi11}c show how this channel can be represented as a PMM. In
this example the PMM has five states $s\in \{0,1,..4\}$, a single
input $x=V$ (the membrane potential) and a single output $y=I$
(the ion current).

Using the Goldman-Hodgkin-Katz (GHK) current equation we have the
following expression for the output function $\omega(x,s)$.
\begin{equation} \label{eq3.7}
   I_j=\omega(V,j)=\frac{p_jz^2FV^{\prime}(C^{in}-C^{out}e^{-zV^\prime}) }{1-e^{-zV^\prime}}
\end{equation}

where
\begin{itemize}
\item $I_j$ is the ion current in state $s=j$ with input $x=V$
\item $p_j\ [cm/sec]$ is the permeability of the channel in state
$s=j$ \item $z$ is the valence of the ion ($z=1$ for $K^+$ and
$Na^+$, $z=2$ for $Ca^{++}$) \item $F=9.6484\cdot10^4\ [C/mol]$ is
the Faraday constant \item $V^{\prime}=\frac{VF}{RT}$ is the ratio
of membrane potential to the thermodynamic potential, where $T\
[K]$ is the absolute temperature, and $R=8.3144\ [J/K \cdot~mol]$
is the gas constant \item $C^{in}$ and $C^{out}\ [mol]$ are the
cytoplasmic and extracellular concentrations of the ion,
respectively
\end{itemize}

One can make different assumptions about the function
$\alpha(x,s_j,s_i)$, describing the conditional probability
densities of state transitions. It is convenient to represent this
function as a matrix of voltage dependent coefficients
$a_{ij}(V)$.
\begin{equation} \label{eq3.9}
   \alpha =
   \begin{pmatrix}
                 &a_{00}(V)\ ..\ a_{0j}(V)\ ..\  a_{0m}(V)\\
                 \\
                 &a_{i0}(V)\ ..\ a_{ij}(V)\ ..\  a_{im}(V)\\
                 \\
                 &a_{m0}(V)\ ..\ a_{mj}(V)\ ..\  a_{mm}(V)
   \end{pmatrix}
\end{equation}
where $m=n-1$. Note that the diagonal elements of this matrix are
not used in equation (\ref{eq3.1}).

In the model of spike generation discussed in Eliashberg (1990a)
both sodium, $Na^+$, and potassium, $K^+$ channels were treated as
PMMs with five states shown in Figure \ref{fi11}. Coefficients
$a_{10}$, $a_{21}$, $a_{32}$ where assumed to be sigmoid functions
of membrane potential, and coefficients $a_{43}$ and $a_{04}$ -
constant. In the case of the sodium channel, $s=3$ was used as a
high permeability state, and $s=4$ was used as inactive state. In
the case of potassium channel, $s=3$ and $s=4$ were assumed to  be
high permeability states.

\subsection{Concept of an Ensemble of Protein Molecule Machines
 (EPMM)} \label{sec3.3}
\textbf{DEFINITION}. An \emph{Ensemble of Protein Molecule
Machines} (EPMM) is a set of identical independent PMMs with the
same input vector, and the output vector equal to the sum of
output vectors of individual PMMs.
\begin{figure}[b!]
\begin{center}
\includegraphics[width=2.8in]{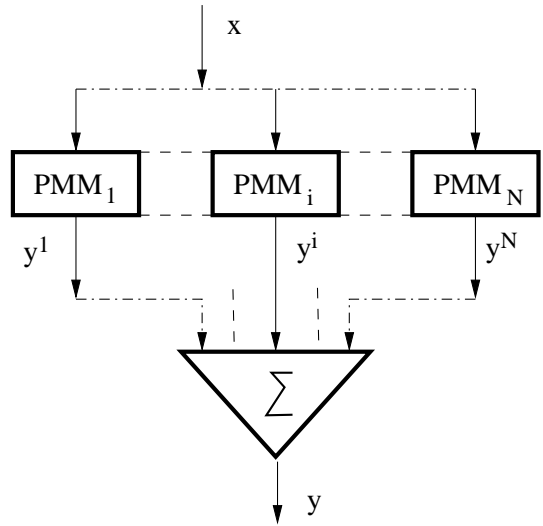}
\end{center}
\caption{The structure of EPMM } \label{fi12}
\end{figure}
\begin{figure}[b!]
\begin{center}
\includegraphics[width=4.2in]{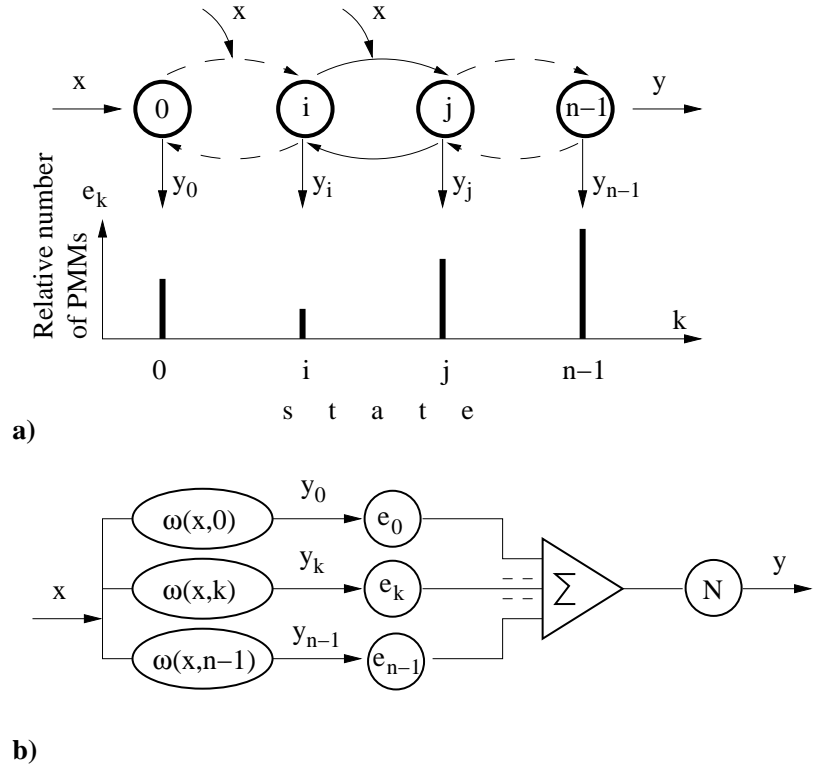}
\end{center}
\caption{E-states as the numbers of PMM's in different states}
\label{fi13}
\end{figure}
The structure of an EPMM is shown in Figure \ref{fi12}, where $N$
is the total number of PMMs, $y^k$ is the output vector of the
$k$-th PMM, and $y$ is the output vector of the EPMM. We have

\begin{equation} \label{eq3.10}
   y=\sum_{k=1}^{N}y^k
\end{equation}

Let $N_i$ denote the number of PMMs in state $s=i$ (the occupation
number of state $i$). Instead of (\ref{eq3.10}) we can write

\begin{equation} \label{eq3.11}
   y = \sum_{i=0}^{n-1}{N_i\omega(x,s_i)}
\end{equation}

\noindent $N_i$ $(i=0,...n-1)$ are random variables with the
binomial probability distributions
\begin{equation} \label{eq3.12}
   P \{ N_i=m \}=\binom{m}{N}P_i^m(1-P_i)^{N-m}
\end{equation}
$N_i$ has the mean $\mu_i=NP_i$ and the variance
$\sigma_i^2=~NP_i(1-P_i)$.

\medskip
\noindent Let us define the relative number of PMMs in state $s=i$
(the relative occupation number of state $i$) as

\begin{equation} \label{eq3.13}
   e_i=\frac{N_i}{N}
\end{equation}

The behavior of the average $\overline{e}_i$ is described by the
equations similar to (\ref{eq3.1}) and~(\ref{eq3.2}).
\begin{align}
   &\frac{d\overline{e}_i}{dt} = \sum_{j\neq i}{\alpha(x,s_i,s_j)\overline{e}_j } - \overline{e}_i \sum_{j\neq i}{\alpha(x,s_j,s_i)}
   \label{eq3.14}\\
   &at\  t=0 \qquad \sum_{i=0}^{n-1}{\overline{e}_i}=1
   \label{eq15}
   \end{align}
\noindent The average output $\overline{y}$ is equal to the sum of
average outputs for all states.

\begin{equation} \label{eq16}
   \overline{y}= N\sum_{i=0}^{n-1}{\omega(x,s_i)\overline{e}_i}
\end{equation}

\noindent The standard deviation for $e_k$ is equal to
\begin{equation} \label{eq17}
   \sigma_k = \sqrt{P_k(1-P_k)/N}
\end{equation}

Figure \ref{fi13} illustrates the implementation of E-states as
relative occupation numbers of the  states of a PMM. The maximum
number of independent E-state variables is equal to $n-1$. The
number is reduced by one because of the additional equation
(\ref{eq15}).

\subsection{EPMM as a Robust Mixed-Signal Computer} \label{sec3.4}

\begin{figure}[b!]
\begin{center}
 \includegraphics[width=3.7in]{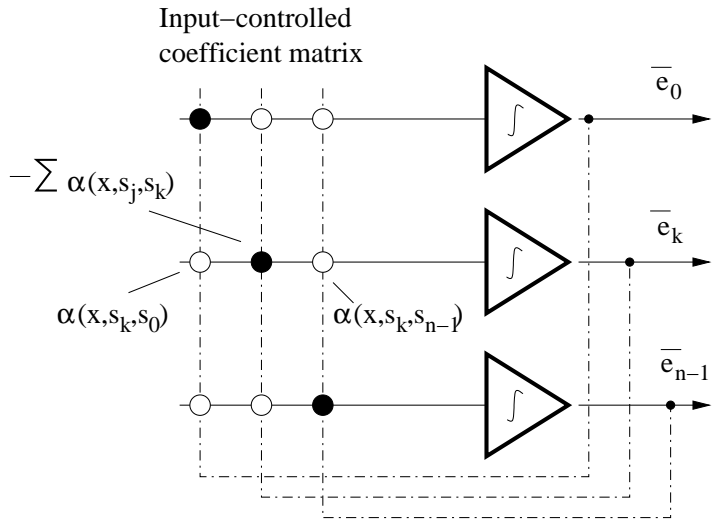}
 \end{center}
 \caption{EPMM as an analog computer
 with an input--controlled coefficient matrix}\label{fi14}
\end{figure}

An EPMM can serve as a robust \emph{analog  computer} with the
{\emph{input--controlled coefficient matrix}} shown in Figure
\ref{fi14}. Because some coefficients in this matrix can change
sharply (almost step-wise) as functions of inputs (e.g., the
membrane potential), an EPMM can be better characterized as a
\emph{mixed-signal} computer. The statistical molecular
implementation of this computer is extremely robust, since all the
characteristics of  the whole computer are determined by the
properties of a single PMM.

It is interesting to emphasize that the matrix of input dependent
coefficients is implemented as the matrix of input dependent
probabilities, so no external connections are needed. It would be
very difficult (if at all possible) to reach this level of
microminiaturization and this level of reliability using
traditional VLSI techniques.

\subsection{On conformational dynamics and chemical kinetics}\label{sec3.5}

When a neural modeler needs to simulate different effects of
cellular STM, he/she usually assumes that these effects are
associated with chemical kinetics and/or with the accumulation of
different neurotransmitters and/or ions in different cellular
compartments. This approach to cellular STM encounters serious
problems:
\begin{enumerate}
\item It is difficult to justify sufficiently big time constants
-- chemical kinetics is quite fast, and cellular compartments are
very small. \item It is difficult to justify  nontrivial
nonlinearities. For example, it is difficult to get different time
constants for increase (charge) and decrease (discharge) of an STM
variable. \item It is difficult (if not impossible) to get
nontrivial timing effects, e.g., different results for different
order of input events.
\end{enumerate}
All these possibilities are readily available with the EPMM
formalism that deals with sophisticated \emph{conformational
dynamics} rather than with a relatively simple \emph{chemical kinetics}.\\\\
\textbf{IMPORTANT}! To avoid common misunderstanding, I want to
emphasize that \emph{conformational dynamics} has nothing to do
with traditional \emph{chemical kinetics}. Conformational dynamics
is determined by the biophysical properties of protein molecules.
No chemistry is involved, for example, in the case of voltage
controlled channels. Even in the case of ligand controlled
channels or enzymes it is inadequate to think about the
interaction between a neurotransmitter molecule and a protein
molecule as a chemical reaction. Protein molecules are very big
($>$ 50,000 Dalton), whereas neurotransmitter molecules are tiny
($<$100 Dalton). A tiny molecule changes the conformation of a big
molecule, so the latter can temporarily open its pore (as in the
case of an ion channel) or become a catalyst producing a second
messenger. (See, for example, Changeux, 1993, and Hille, 2001.)

\subsection{What can be computed with EPMM's? }
\label{sec3.6}
\begin{figure}[t!]
\begin{center}
 \includegraphics[width=3.3in]{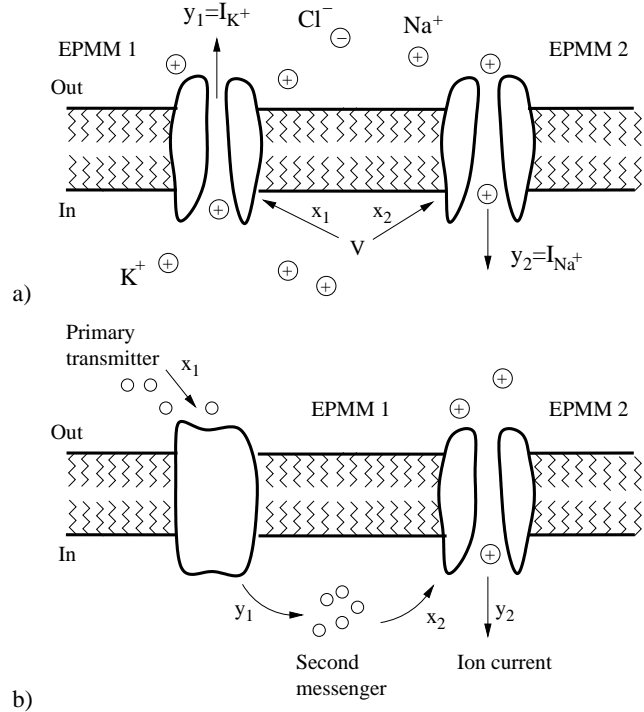}
 \end{center}
 \caption{Two EPMMs interacting via (a) electrical and (b) chemical messages}\label{fi15}
\end{figure}
Very little is known about the properties of different membrane
proteins to represent them as abstract probabilistic nanomachines.
The best studied are the sodium and potassium channels used in the
classical Hodgkin and Huxley (1952) model for the generation of
nerve spike. It is believed that these protein molecules have
close to five different states each. In this specific case, the
EPPM formalism gives a good approximation of the available
experimental data (Eliashberg, 1990a, 2003). Therefore, it seems
reasonable to believe that this formalism should work well in many
other less studied cases.

A single neuron can have several different EPMMs interacting via
electrical messages (membrane potential) and chemical messages
(different kinds of neurotransmitters). As  mentioned in Section
\ref{sec3.2}, the Hodgkin-Huxley (1952) model can be naturally
expressed in terms of two  EPMMs (corresponding to the sodium and
potassium channels) interacting via common membrane potential (see
Figure \ref{fi15}a).  Figure \ref{fi15}b shows two EPMMs
interacting via a second messenger. In this example, EPMM1 is the
primary transmitter receptor and EPMM2 is the second messenger
receptor.

Some examples illustrating nontrivial computational possibilities
of the EPMM formalism will be discussed in Part II of this
paper.
\subsection{The main statements} \label{sec3.7}
\begin{enumerate}
\item The whole human brain is a nonclassical symbolic system --
an E-machine (Eliashberg, 1967, 1979). \item The popular notion
that the brain implements multidimensional real mapping is a
fallacy. The whole concept of a learning algorithm that optimizes
synaptic weights to create the above mappings is largely
irrelevant to the problem of human learning. \item The main data
storage procedure of the human brain must be universal -- close to
"memorizing raw experience." Instead of processing data before
placing it in memory, the brain must process "raw" data
dynamically (on the fly) depending on context. No
context-dependent statistics can be precalculated in advance, in
principle, because the number of possible contexts explodes
combinatorially. \item Biological neural networks have the right
computational resources to implement the above dynamic approach.
The main computational engine of the brain is the statistical
mechanics of protein nanomachines rather than the "statistical
mechanics of neural networks." The notion of a neuron as a simple
atomic computing element, employed by the latter approach, is
inconsistent with the available neurobiological data (Kandel and
Spenser, 1968; Kandel, Jessel, and Schwartz, 2000; Nichols,
Martin, Wallace, 1992, Byrne, 1987).
\end{enumerate}

\textbf{REFERENCES}\\\\
Anderson, J.R. (1976). Language, Memory, and Thought. \emph{Hillsdale,
New Jersey: Lawrence Erlbaum Associates, Publishers}.\\\\
Baddeley, A.D. (1982). Your memory: A user's
guide. \emph {MacMillan Publishing Co., Inc.}\\\\
Burns, B.D., (1958). The Mamalian Cerebral Cortex.
\emph{Arnold}.\\\\
Byrne, J.H. (1987). Cellular Analysis of Associative Learning.
\emph{Physiological Review}, 67, 2, 329-439.\\\\
Changeux, F. (1993). Chemical Signaling in the Brain.
\emph{Scientific American, November}, 58-62.\\\\
Chomsky, N. (1956). Three models for the description of language.
\emph{I.R.E. Transactions on Information Theory. JT-2},
113-124.\\\\
Collins, A.M., and Quillian, M.R., (1972). How to make a language
user. \emph{In E. Tulving and W. Donaldson (Eds.) Organization and
memory. New York: Academic Press}.\\\\
Eliashberg, V. 1967. On a class of learning machines. \emph{Moscow: Proceedings of VNIIB, \#54}, 350-398.\\\\
Eliashberg, V. (1979). The concept of E-machine and the problem of
context-dependent behavior. \emph{TXU 40-320, US Copyright
Office}.\\\\
Eliashberg, V. (1981). The concept of E-machine: On brain hardware
and the algorithms of thinking. \emph{Proceedings of of the Third
Annual Meeting of Cognitive Science Soc.}, 289-291.\\\\
Eliashberg, V. (1989). Context-sensitive associative memory:
"Residual excitation" in neural networks as the mechanism of STM
and mental set. \emph{Proceedings of IJCNN-89, June 18-22, 1989,
Washington, D.C}. vol. I, 67-75.\\\\
Eliashberg, V. (1990a). Molecular dynamics of short-term memory.
\emph{Mathematical and Computer modeling in Science and
Technology}. vol. 14, 295-299.\\\\
Eliashberg, V. (1990b). Universal learning neurocomputers.
\emph{Proceeding of the Fourth Annual parallel processing
symposium. California state university, Fullerton}. April 4-6,
1990., 181-191.\\\\
Eliashberg, V. (1993). A relationship between neural networks and
programmable logic arrays. \emph{International Conference on
Neural Networks, San Francisco, CA.} March 28- April 1, 1993.
0-7803-0999-5/93, IEEE, 1333-1337.\\\\
Eliashberg, V. (2002). What Is Working Memory and Mental Imagery?
A Robot that Learns to Perform Mental Computations. \emph{Web
publication. (Available at www.brain0.com and
http://arxiv.org/abs/cs.AI/0309009. )}\\\\
Eliashberg, V. (2003). Ensembles of protein molecules as
statistical analog computers. In press. (Available at
http://arxiv.org/abs/physics/030804 and at www.brain0.com)\\\\
Hille, B. (2001). Ion Channels of Excitable Membranes.
\emph{Sinauer Associates. Sunderland}, MA\\\\
Hodgkin, A.L., \& Huxley, A.F. (1952). A quantitative description
of membrane current and its application to conduction and
excitation in nerve. \emph{Journal of Physiology, 117}, 500-544.\\\\
Kandel, E.R., and Spencer, W.A. (1968). Cellular
Neurophysiological Approaches in the Study of Learning.
\emph{Physiological Rev.} 48, 65-134.\\\\
Kandel, E., Jessel,T., Schwartz, J. (2000). Principles of Neural
Science. \emph{McGraw-Hill}.\\\\
Meynert, T. (1884). Psychiatrie. Wien.\\\\
Miller, G.A. (1956). The magical number seven, plus or minus two:
Some limits on our capacity for processing information.
\emph{Psychological Review, 63}. 81-97.\\\\
Minsky, M.L. (1967). Computation: Finite and Infinite Machines,
\emph{Prentice-Hall, Inc}.\\\\
Nichols, J.G., Martin, A.R., Wallace B.G., (1992) From Neuron to
Brain, \emph{Third Edition, Sinauer Associates}.\\\\
Pinker, S., Mehler, J. (Eds.) (1988). Connections and Symbols. \emph{The
MIT Press, Cambridge}.\\\\
Rosenblatt, F. (1962). Principles of neurodynamics. Perceptron and
the Theory of Brain Mechanisms. \emph{Spartan Books. Washington
D.C}.\\\\
Rumelhart, D.E., McClelland, J.L. (Eds.) (1986). Parallel
Distributed processing: Explorations in the Microstructure of
Cognition. \emph{Cambridge, MA: MIT Press.} (Vols 1 and 2.)\\\\
Sperling, G.A. (1960). The information avalable in brief
presentations. \emph{Psychological Monographs, 74, No. 498}.\\\\
Steinbuch, K. (1961). Die Lernmatrix. \emph{Kybernetik }1,
36-45.\\\\
Turing, A.M. (1936). On computable numbers, with an application to
the Entscheidungsproblem. \emph{Proc. London Math. Society}, ser.
2, 42\\\\
Varju D. (1965). On the Theory of Lateral Inhibition, Consiglio
Nazionalle Delle Reicerche Quardeni de "La Ricerca Scientifica,"
v. 31 \\\\
Vvedensky, N.E. 1901. Excitation, inhibition and narcosis.
\emph{In "Complete collection of works"}. USSR, 1953.\\\\
Widrow, B. (1962). Generalization and information storage in
networks of Adaline neurons. \emph{Self-organizing systems.
Washington DC: Spartan Books, Cambridge MA: MIT Press}.\\\\
Zopf, G.W. (1962). Attitude and Context. In "Principles of
Self--organization". \emph{Pergamon Press}, 325-346.

\end{document}